%% file: acl_latex.tex
\documentclass[11pt]{article}

\usepackage[final]{acl}
\usepackage{microtype}
\usepackage{hyperref}
\usepackage{url}
\usepackage{booktabs}
\usepackage{tikz}
\usepackage{xcolor}
\usepackage{lipsum} 
\usepackage{wrapfig}
\usepackage[utf8]{inputenc}
\usepackage{graphicx}
\usepackage{placeins}
\usepackage[most]{tcolorbox}
\usepackage{lipsum}
\usepackage{caption}
\usepackage{microtype}
\usepackage{listings} 
\usepackage{inconsolata}
\usepackage{import}
\usepackage{microtype}
\usepackage{layout}
\usepackage{tabularx, makecell}
\usepackage{booktabs}
\usepackage{mathrsfs}
\usepackage{amssymb} 
\usepackage{url}
\usepackage{graphicx}
\usepackage{xspace,paralist}
\usepackage{amsmath}
\usepackage{appendix}
\usepackage{comment} 
\usepackage{makecell}
\usepackage{multirow}
\usepackage{colortbl}
\usepackage{xcolor}
\usepackage{tablefootnote}
\usepackage{longtable}
\usepackage{lipsum}
\usepackage{xspace}
\usepackage{bbding}
\usepackage{pifont}
\usepackage{arydshln}
\usepackage{array}
\usepackage{cleveref}
\usepackage{hyperref}
\usepackage{subcaption}
\usepackage{multirow}
\usepackage{arydshln}
\usepackage{scalerel}
\usepackage[russian,english]{babel}
\usepackage{CJKutf8}
\usepackage{footmisc}
\usepackage{placeins}
\newcolumntype{L}[1]{>{\raggedright\arraybackslash}p{#1}}

\usepackage[T2A]{fontenc}
\usepackage[russian,english]{babel}

\definecolor{darkgreen}{rgb}{0.0, 0.5, 0.0}

\usepackage{CJKutf8}

\usepackage{float}
\restylefloat{table}

\usepackage[T1]{fontenc}
\usepackage{arydshln}

\usepackage{lineno}

\usepackage{times}
\usepackage{latexsym}

\usepackage[T1]{fontenc}

\usepackage[utf8]{inputenc}

\usepackage{microtype}

\usepackage{inconsolata}

\usepackage{graphicx}

\title{Exploring Language-Agnosticity in Function Vectors:\\
A Case Study in Machine Translation}

\author{
Nurkhan Laiyk$^{\heartsuit}$\thanks{Correspondence to: \texttt{nurkhan.laiyk@mbzuai.ac.ae}.},  Gerard I. Gallego$^{\diamondsuit\P}$, Javier Ferrando\thanks{This work is not related to the author's position at Amazon.}, Fajri Koto$^{\heartsuit}$ \\
$^{\heartsuit}$Mohamed bin Zayed University of Artificial Intelligence \\
$^{\diamondsuit}$Universitat Politècnica de Catalunya \quad \\
$^{\P}$Cantina Labs}

\begin{document}
\maketitle

\begin{abstract}
Function vectors (FVs) are vector representations of tasks extracted from model activations during in-context learning. While prior work has shown that multilingual model representations can be language-agnostic, 
it remains unclear whether the same holds for function vectors. We study whether FVs exhibit language-agnosticity, using machine translation as a case study. Across three decoder-only multilingual LLMs, we find that translation FVs extracted from a single English$\to$X direction transfer to other target languages, consistently improving the rank of correct translation tokens across multiple unseen languages. We further find that the highest-gain tokens span multiple languages and that translation FVs across directions share most of their top-ranked heads, indicating that the FV encodes a largely language-agnostic translation signal rather than a language-pair-specific mapping. 
\end{abstract}

\input{sections/1_introduction}

\input{sections/2_related_work}

\input{sections/3_methodology}
\input{sections/4_experiments}

\input{sections/5_conclusion}
\input{sections/6_limitations}
\bibliography{custom}

\appendix
\input{sections/7_appendix}

\end{document}

%% file: sections/1_introduction.tex
\section{Introduction}

Large language models (LLMs) have become central to multilingual NLP, showing strong performance across a wide range of cross-lingual and multilingual tasks~\citep{brown2020language}. A growing body of mechanistic work has begun to ask how these models represent and process information across languages~\citep{resck-etal-2025-explainability,ferrando2024primerinnerworkingstransformerbased}. One recurring finding is that many internal representations are \emph{language-agnostic}: they encode task-relevant content, such as semantic meaning, 
independent of language identity, so that similar concepts in different languages are mapped to nearby states in activation space.

 \begin{figure}[t]
    \centering
    \includegraphics[width=0.95\linewidth]{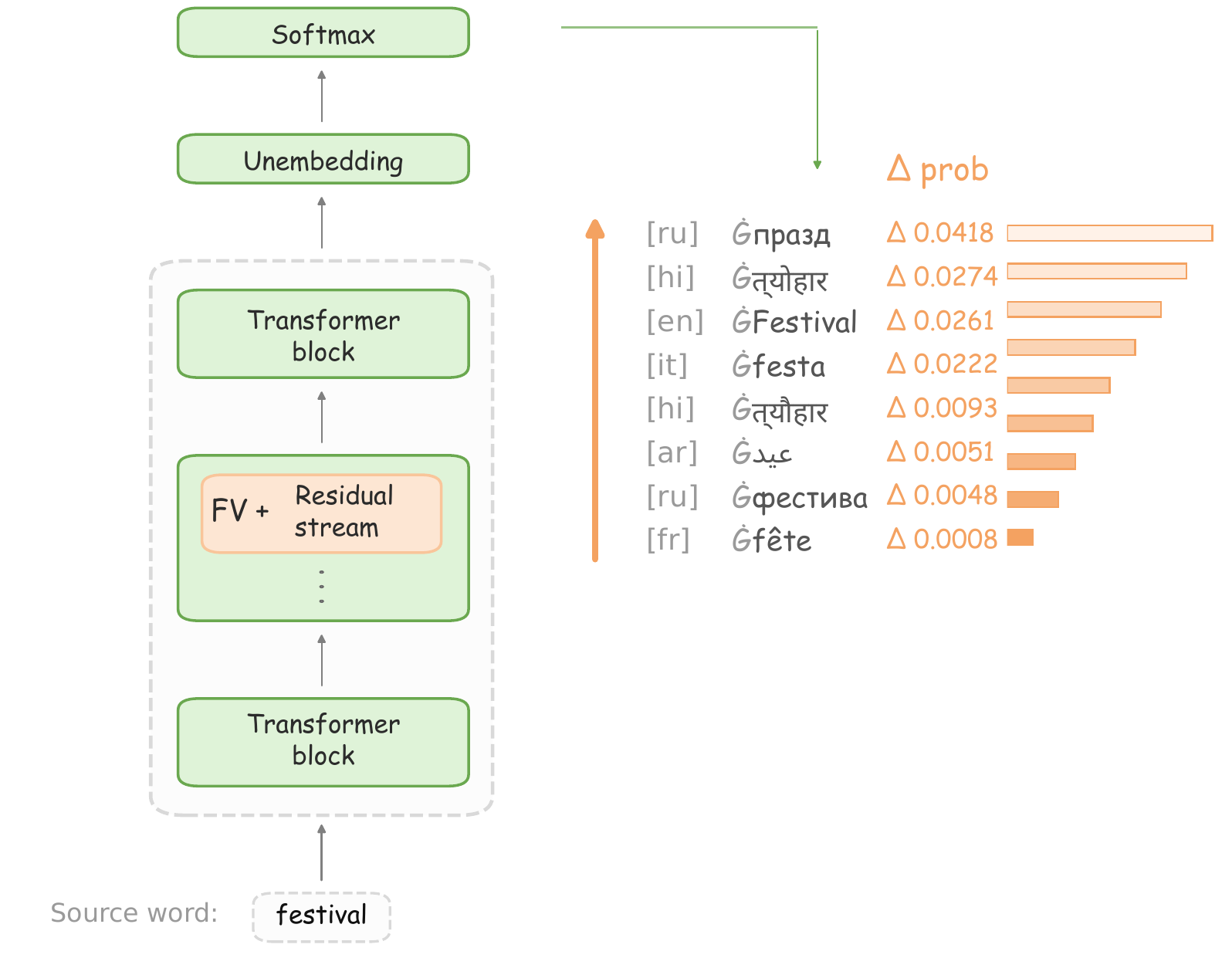}
    \caption{\textbf{Example of FV effects on the next-token distribution.} For a single source word, we show the top tokens whose probability ($\Delta p$) increases the most after FV, highlighting that the largest gains often correspond to plausible translations across multiple languages.}
    \label{fig:top-tokens-gain-intro}
\end{figure}
This property has been documented at several levels of abstraction. At the representation level, work on multilingual BERT \cite{devlin-etal-2019-bert} showed that substantial cross-lingual alignment can emerge without explicit parallel supervision, enabling zero-shot transfer across languages and scripts~\citep{pires-etal-2019-multilingual, wu-dredze-2019-beto}. More recent studies on autoregressive LLMs have confirmed that concept representations can be largely language-agnostic, with intermediate activations partially separating ``what is being said'' from ``in which language'' it is expressed~\citep{wendler-etal-2024-llamas, dumas-etal-2025-separating}. Beyond static representations, language-agnosticity has also been observed at the level of grammatical concepts~\citep{brinkmann-etal-2025-large} and at the level of circuits, where prior work suggests that some internal mechanisms are shared across languages~\citep{ferrando-costa-jussa-2024-similarity,zhang2025the}.

Building on this line of work, we ask whether \textit{function vectors} exhibit the same language-agnostic property. Function vectors (FVs), introduced by \citet{todd2024function}, are compact additive representations of tasks extracted from in-context learning demonstrations. Recent work has shown them to be a central mechanism behind in-context learning \cite{yin2025which} and a useful analytical lens on model behavior more broadly \cite{jiang2025unlocking}, making them a natural object through which to study how multilingual capabilities are internally represented. Using machine translation as a case study, we extract translation FVs from English$\to$X word-level prompts in three multilingual decoder-only LLMs and apply them as residual-stream interventions. We find that FVs extracted from a single English$\to$X direction promote correct translations across multiple target languages (Figure~\ref{fig:top-tokens-gain-intro}), and we confirm the cross-lingual nature of this effect through high-gain token analysis and substantial overlap in the selected FV heads across translation directions.

%% file: sections/2_related_work.tex
\section{Related work}



Earlier work on understanding translation in neural models focused on observational analyses such as neuron activation patterns and representation probing~\citep{mu2024faithful, tang2024language, wendler-etal-2024-llamas}. More recent studies have moved toward causal, circuit-level analyses that identify the specific components responsible for translation behavior. \citet{zhang2025exploring} study word-level translation in decoder-only LLMs and find that the behavior depends on a sparse subset of attention heads and MLPs with distinct functional roles, and that fine-tuning these components improves translation while preserving general capabilities. \citet{lasnier2026disentangling} extend this to sentence-level few-shot MT, using activation patching to decompose translation into two separable subtasks, target language identification and sentence equivalence, each implemented by sparse attention-head sets, and use the identified heads to build steering vectors for instruction-free MT.

%% file: sections/3_methodology.tex
\begin{figure*}[h]
    \centering
    \begin{minipage}[t]{0.4\textwidth}
        \centering
        \includegraphics[width=\linewidth]{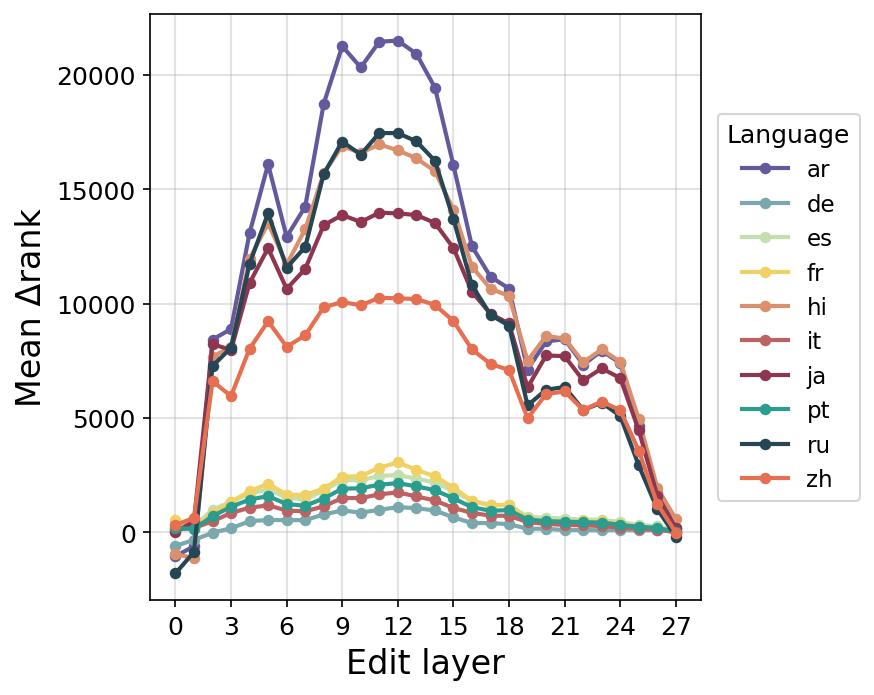}
    \end{minipage}
    \hfill
    \begin{minipage}[t]{0.55\textwidth}
        \centering
        \includegraphics[width=\linewidth]{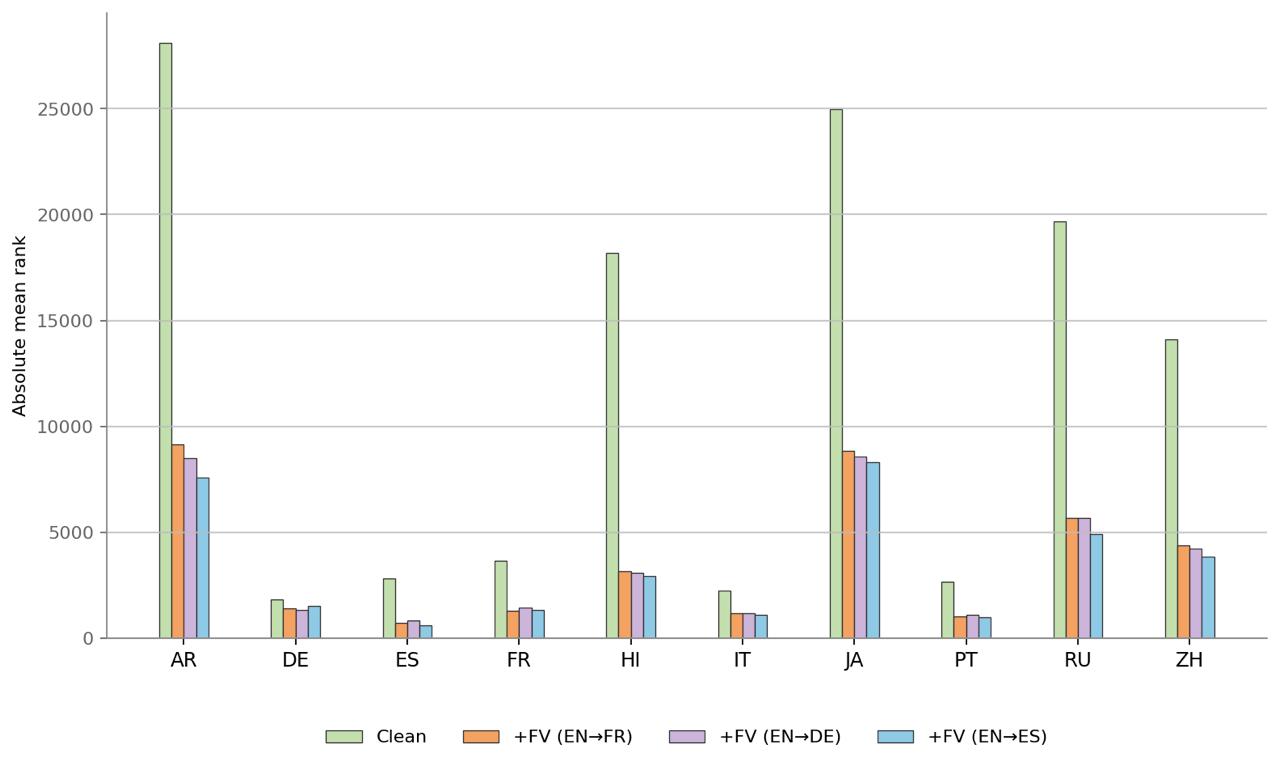}
    \end{minipage}
    \caption{Left: average mean $\Delta$rank across edit layers for different target languages for Llama-3.2-3B. Curves correspond to target languages, and higher values mean that the FV moves the correct translation token closer to the top of the next-token distribution. 
    Right: per-language absolute mean rank for LLaMA-3.2-3B under the clean setting and after FV intervention. }
    \label{fig:avg-mean-rank-llama}

\end{figure*}

\begin{figure}[t]
    \centering
    \includegraphics[width=\linewidth]{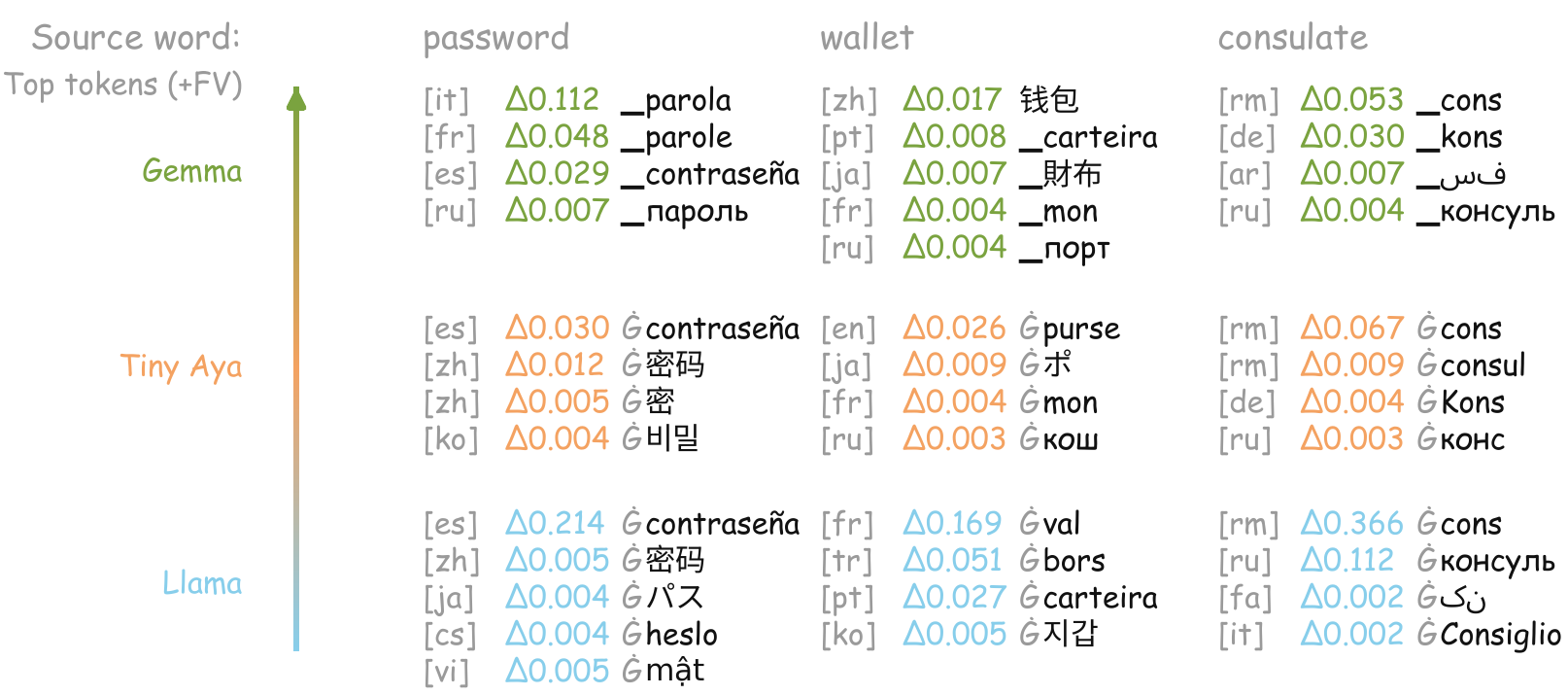}
    \caption{\textbf{Token-level effect of FV steering.}
    For source words, we compare the next-token distribution with and without FV injection and list tokens ranked by their probability increase under the intervention (\(+\)FV).}
    \label{fig:top-tokens-gain}
\end{figure}
\section{Methodology}
\label{sec:background}

\paragraph{Function vectors}\label{para:background} We briefly review the function vector framework of \citet{todd2024function}, on which our analysis builds. 
For each task $t$, we assume a dataset of ICL prompts $P_t$ and a corrupted version $\tilde P_t$ formed by shuffling demonstration labels. The procedure first selects a small set of attention heads using activation patching. For a given task $t$, it compares prompts from $P_t$ (correct demonstrations) to corrupted prompts $\tilde p \in \tilde P_t$. For each head $(\ell,j)$, the head output on a corrupted prompt $\tilde p$ is patched with the head's mean output over $p \in P_t$, denoted $\bar a^t_{\ell j}$. The causal indirect effect (CIE) of head $(\ell,j)$ is defined as the resulting change in the probability of the correct answer token $y$:
\[
\mathrm{CIE}_{\ell j}(\tilde p)
=
P\!\left(y \mid \tilde p;\; a_{\ell j}\leftarrow \bar a^t_{\ell j}\right)
-
P\!\left(y \mid \tilde p\right).
\]
Heads are ranked by their average indirect effect across corrupted prompts, and the top heads form the selected set $A$. The function vector for task $t$ is then defined as the sum of the task-conditioned mean outputs of the selected heads,
\[
v_t=\sum_{(\ell,j)\in A}\bar a^t_{\ell j},
\]
and is applied by adding $v_t$ to the residual stream at a chosen layer during inference. In our experiments, we apply a norm-matched variant, $h'_\ell = h_\ell + \alpha \|h_\ell\| \hat{v}_t$ with $\hat{v}_t = v_t / \|v_t\|$, to account for the growth of residual norms with depth.

\paragraph{Extracting FV for Machine Translation}
To clarify how the general FV framework applies to translation, we now detail the structure of the ICL prompts and the corruption scheme specific to word-level machine translation. Each ICL prompt $p^t_i$ consists of $N$ demonstration pairs followed by a single query word:
\[
p^t_i = \bigl[(x_{i1},\, y_{i1}),\; (x_{i2},\, y_{i2}),\; \dots,\; (x_{iN},\, y_{iN}),\; x_{iq}\bigr],
\]
where each $x_{ik}$ is a word in the source language (e.g., English) and $y_{ik}$ is its translation in the target language (e.g., French). For instance, consider an English$\to$French prompt of the following form:

\begin{center}
\texttt{Q: eggs\textbackslash nA: oeufs\textbackslash n\textbackslash nQ: good\textbackslash nA: bon\textbackslash n\textbackslash nQ: sugar\textbackslash nA: sucre\textbackslash n\textbackslash nQ: loan\textbackslash nA:}
\end{center}

\noindent Given this prompt, the model must infer the translation mapping from the demonstrated pairs and produce \texttt{pr\^{e}t}, the correct French translation of the query word \texttt{loan}. 
The correct answer token $y$ used in the FV scoring procedure is 
the gold target token.\footnote{If the gold translation tokenizes into multiple subwords, we use only its first token.} To form the corrupted prompts $\tilde{p} \in \tilde{P}_t$, the source-side words and the query remain fixed while the target words $\{y_{i1},\dots,y_{iN}\}$ are randomly permuted across the $N$ demonstrations, breaking the source--target mapping. The FV extraction then proceeds as described in Paragraph~\ref{para:background}.

\paragraph{Models and Data}
\label{para:models-data}
We focus on decoder-only autoregressive LLMs and evaluate three models: Gemma-2-2B \citep{riviere2024gemma2}, Llama-3.2-3B \citep{grattafiori2024llama3,llama32card}, and Tiny Aya \citep{tinyaya2026}.
We choose these models because they provide multilingual coverage at small parameter scales, which enables systematic FV extraction and intervention sweeps within our computational budget.
We use the English--French, English--German, and English--Spanish word-pair datasets from \citet{todd2024function} to compute the translation FVs, with 4705, 5154, and 5200 pairs, respectively.

%% file: sections/4_experiments.tex
\section{Experiments and Results}

\paragraph{Cross-lingual rank improvement.}
We evaluate whether the translation FV exhibits language-agnostic transfer by promoting correct translations across target languages in the next-token distribution. Following the standard FV evaluation setup, we use zero-shot prompts at evaluation. For each English source word $w_{\text{src}}$ and target language, we collect the gold translation token and measure its rank under the clean model and after FV intervention.\footnotemark[1] 
For evaluation, we build a separate held-out word-to-word set of 120 English source words in Spanish, French, German, Italian, Portuguese, Russian, Arabic, Chinese, Japanese, and Hindi.

We formalize this rank-based evaluation as follows.
Let $M_{\text{clean}}(\cdot \mid w_{\text{src}})$ denote the model’s next-token distribution at the answer position under the clean prompt, and let $M_{\text{FV}}(\cdot \mid w_{\text{src}})$ denote the corresponding distribution under the FV intervention. For each target token $t_{\text{tgt}}$, we measure the change in its rank:

\begin{equation}
\begin{aligned}
\Delta \text{rank}(t_{\text{tgt}} \mid w_{\text{src}})
&= \text{rank}\!\Big(M_{\text{clean}}(t_{\text{tgt}} \mid w_{\text{src}})\Big) \\
&\quad - \text{rank}\!\Big(M_{\text{FV}}(t_{\text{tgt}} \mid w_{\text{src}})\Big).
\end{aligned}
\end{equation}

A positive $\Delta$rank indicates that the FV moves the gold token closer to the top of the distribution. 
Figure~\ref{fig:avg-mean-rank-llama} (left) shows the mean\footnote{per-direction results are reported in Appendix~\ref{sec:llama-three-in-row}} $\Delta$rank per target language, averaged across the three 
experimental conditions (using the EN$\to$ES, EN$\to$DE, and EN$\to$FR FVs separately). The intervention consistently improves the rank of the gold first translation token across target languages, even though each FV is extracted from only a single English$\to$X direction. This suggests that the induced effect is not limited to the target language used to construct the FV, but extends to other target languages as well. A second consistent pattern is that these gains are concentrated in a similar band of middle layers, rather than being distributed uniformly across all layers. This is consistent with prior work that also finds translation-related effects to be strongest in intermediate layers \citep{zhang2025exploring}. Although the magnitude varies across target languages, this middle-layer preference is stable across all three FVs.

One limitation of the $\Delta$rank metric on its own is that it conflates two qualitatively different situations: a small $\Delta$rank may reflect either a weak FV effect or, alternatively, a gold token that was already highly ranked in the clean distribution and therefore has little headroom for improvement. To distinguish these cases, we also examine the absolute mean rank of the gold translation token before and after FV intervention (Figure~\ref{fig:avg-mean-rank-llama}, right).

The results confirm that the cross-lingual effect holds across all evaluated languages, and clarify the role of starting rank. For Arabic, Hindi, Russian, Japanese, and Chinese, the gold token starts from a substantially lower clean rank, and these languages accordingly exhibit the largest absolute improvements under intervention, consistent with the higher $\Delta$rank values in Figure~\ref{fig:avg-mean-rank-llama} (left). For languages where the gold token already ranks favorably in the clean distribution, Spanish, French, German, Italian, and Portuguese, the intervention still shifts it upward, though the room for improvement is inherently smaller. Notably, the FV produces a positive shift in both regimes, indicating that the smaller $\Delta$rank values for these latter languages reflect limited headroom rather than a weak intervention.
\paragraph{Promoted tokens span multiple languages.}
We then manually inspect which tokens gain the most probability under FV intervention. As shown in Figure~\ref{fig:top-tokens-gain}, the highest-gain tokens are plausible translations of the source word and span multiple target languages rather than being limited to the single language used to extract the FV. Additional examples are provided in Appendix~\ref{app:high-token-appendix}. Together with the $\Delta$rank results, this confirms that the FV induces a broadly multilingual translation effect rather than a language-pair-specific mapping.

\paragraph{Top heads overlap across translation directions.}
A natural follow-up question is \emph{why} a single-direction FV promotes translations across languages. One explanation is that the FV relies on attention heads that are themselves shared across translation directions. To test this, we compare the top-ranked heads selected by the FV scoring procedure across several English$\to$X translation FVs. We quantify overlap among the top-10 FV-ranked heads across directions, reporting the number of heads shared across all directions.
The overlap is substantial across all models: 8 of 10 heads are shared in Gemma-2-2B, 7 in Llama-3.2-3B, and 6 in Tiny Aya.
This suggests that part of the FV-associated head set is shared across directions. This finding is consistent with concurrent work that has reported overlap of translation-relevant heads across directions using causal analyses outside the FV framework \citep{zhang2025exploring}.


\paragraph{Ablation and specificity.}

To further confirm that the FV direction carries translation-relevant information rather than an artifact of the addition intervention, we ablate it by projecting it out of the residual stream. Removing the direction degrades the gold-translation rank consistently across all three FVs, and the degradation extends beyond the extraction direction to other target languages, mirroring the cross-lingual pattern observed under FV addition. To assess specificity, we evaluate the ablated model on a suite of standard NLP benchmarks (ARC-Easy~\cite{allenai:arc}, BoolQ~\cite{clark2019boolq}, COPA~\cite{roemmele2011copa}, HellaSwag~\cite{zellers2019hellaswag}, OpenBookQA~\cite{OpenBookQA2018}, PIQA~\cite{Bisk2020}, SciQ~\cite{SciQ}, WinoGrande~\cite{ai2:winogrande}, and MMLU in German, Spanish, and French); accuracy changes by at most 1--2 points across all models and tasks, indicating that the ablation is largely selective to translation. Full results are in Appendix~\ref{app:ablation}.

%% file: sections/5_conclusion.tex
\section{Conclusions}
We investigated whether function vectors exhibit language-agnosticity, using machine translation as a case study. Across three multilingual LLMs, we found that translation FVs extracted from a single English$\to$X direction consistently promote correct translations into multiple target languages, with effects concentrated in a shared band of middle layers. Ablating the FV degrades translation across languages while leaving general task performance intact, confirming that the FV captures a translation-specific subspace. The cross-lingual effect is further supported by substantial overlap in the top-ranked FV heads across translation directions.  Together, these results suggest that function vectors encode a largely language-agnostic translation signal, extending the language-agnosticity property previously observed for representations, grammatical concepts, and circuits. 
 

%% file: sections/6_limitations.tex
\section{Limitations}

Our study has several limitations that should be considered when interpreting the results.

\paragraph{Model scale.}
We evaluate on three models with 2--3B parameters. While this scale enables systematic intervention sweeps, computational constraints prevent us from testing whether the same patterns 
hold at larger scales.

\paragraph{Word-level focus.}
Our experiments operate at the word level, which provides a controlled setting for FV extraction and evaluation but does not capture the full complexity of translation. We do not evaluate on sentence- or document-level translation, nor on more challenging phenomena such as idioms, discourse-level coherence, or morphological agreement. Whether the language-agnostic transfer we observe extends to these settings remains an open question.


\paragraph{Evaluation metric.}
We rely on $\Delta$rank as our primary metric, which measures relative improvement in token ranking but does not directly correspond to translation accuracy or fluency. A token that moves from rank 10{,}000 to rank 5{,}000 shows a large $\Delta$rank but is still far from being the top prediction.

%% file: sections/7_appendix.tex
\newpage
\section{Appendix}

\subsection{Extended high-gain token examples}
We provide additional high-gain token examples in Figures~\ref{fig:high-tokens-gain}, complementing the examples shown in the main text.

\label{app:high-token-appendix}

\begin{figure}[h]
    \centering
    \includegraphics[width=0.85\linewidth]{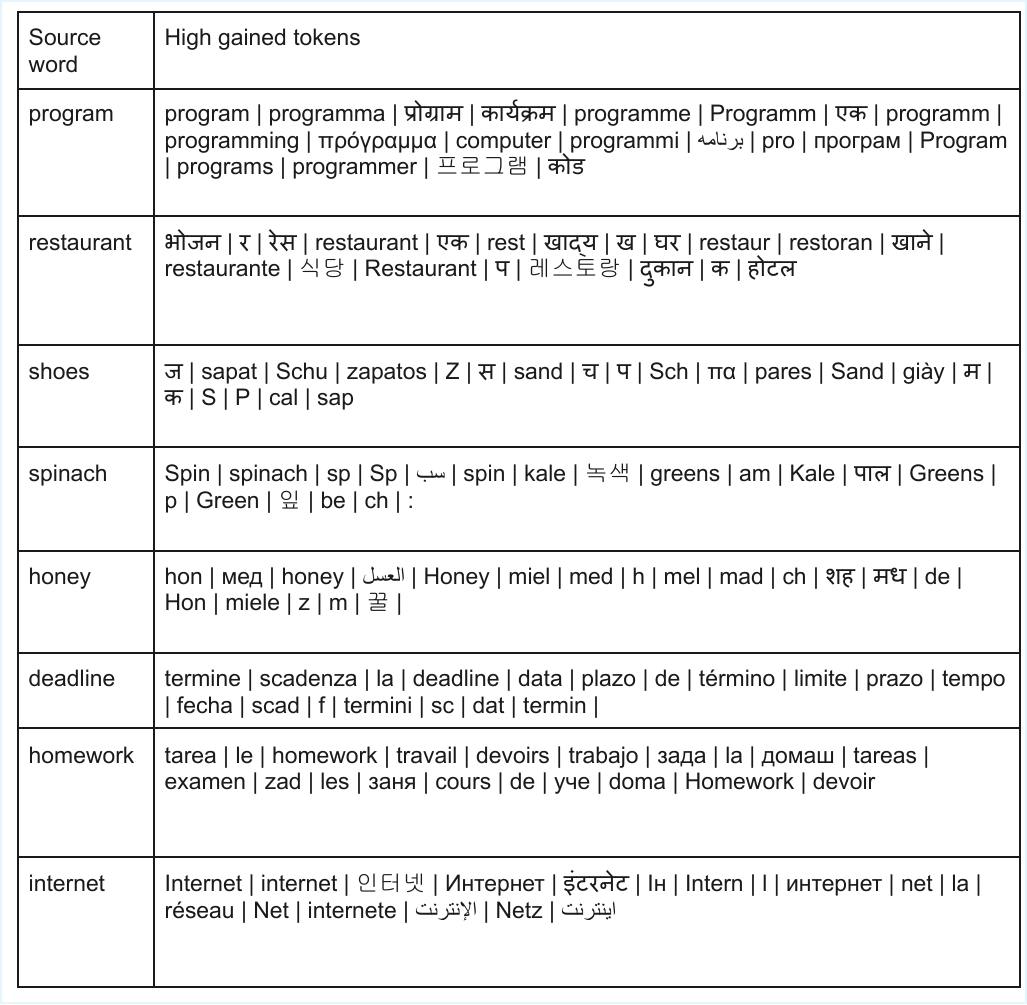}
    \caption{\textbf{Example of FV effects on the next-token distribution for Tiny-Aya.} For a single source word, we show the top tokens whose probability increases the most after FV, highlighting that the largest gains often correspond to plausible translations across multiple languages.}
    \label{fig:high-tokens-gain}
\end{figure}
\FloatBarrier

\begin{figure}[h]
    \centering
    \includegraphics[width=0.85\linewidth]{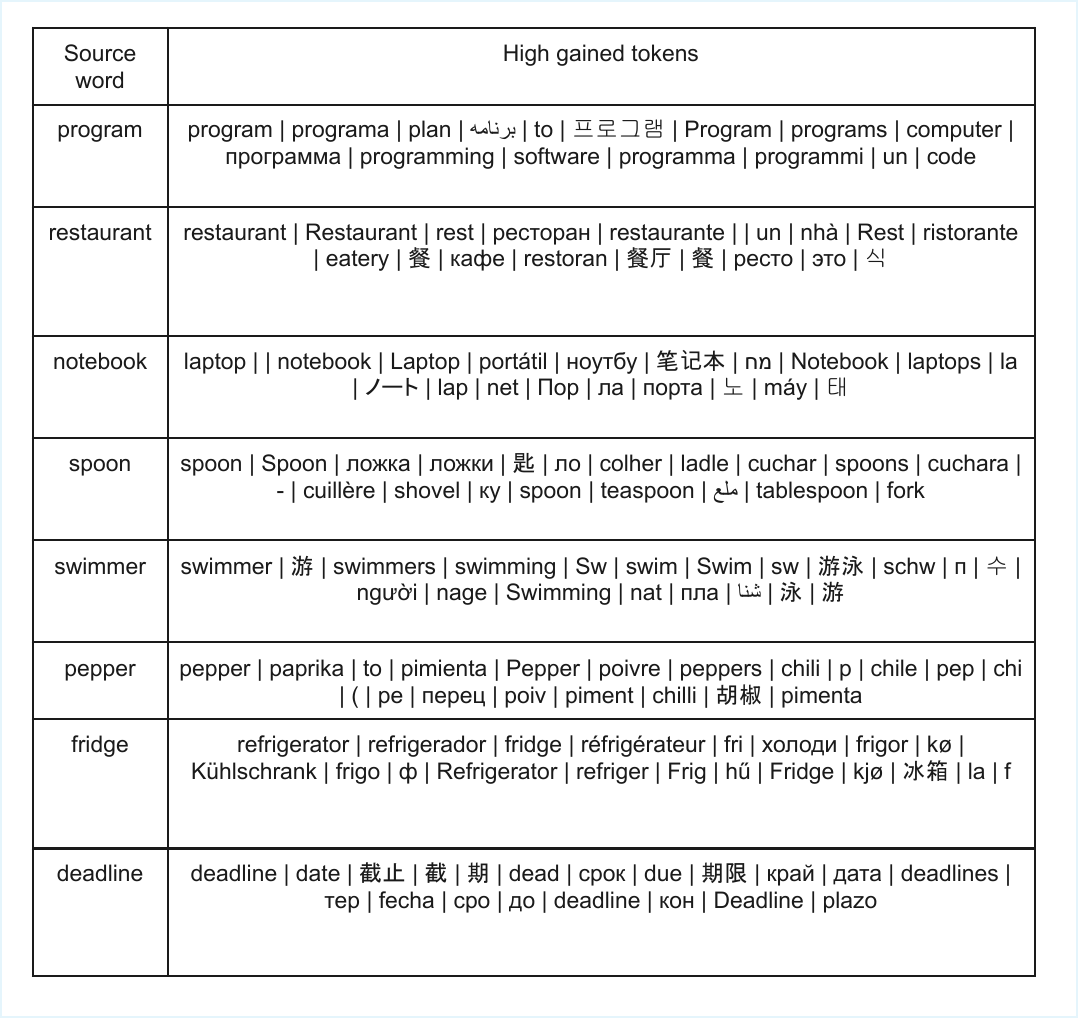}
    \caption{\textbf{Example of FV effects on the next-token distribution for Gemma-2-2B.} For a single source word, we show the top tokens whose probability increases the most after FV, highlighting that the largest gains often correspond to plausible translations across multiple languages.}
    \label{fig:high-tokens-gain}
\end{figure}

\subsection{Per-direction mean $\Delta$rank results}
\begin{figure*}[t]
    \centering
    \includegraphics[width=\linewidth]{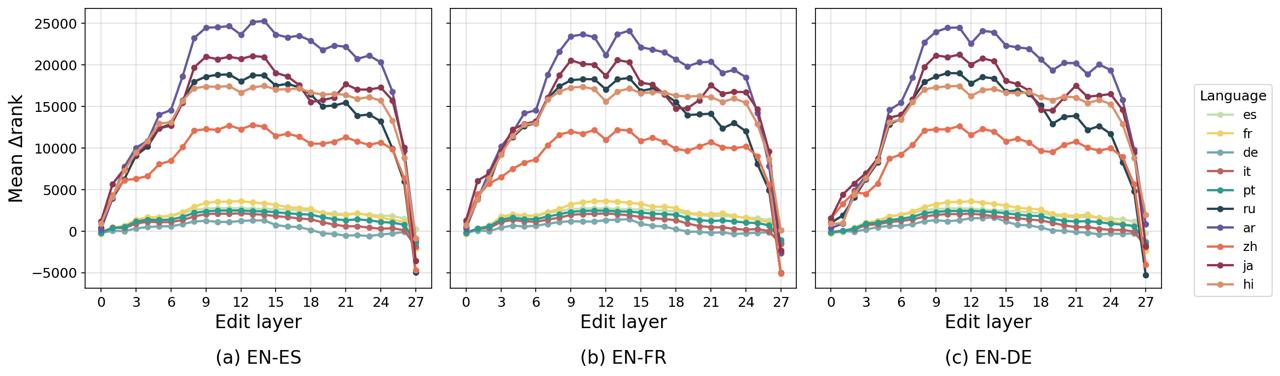}
    \caption{\textbf{Mean $\Delta$rank across edit layers for different target languages for Llama-3.2-3B.} Curves correspond to target languages, and higher values mean that the FV
moves the correct translation token closer to the top of the next-token distribution. 
}
    \label{fig:mean-rank-llama}
\end{figure*}

\begin{figure*}[t]
    \centering
    \includegraphics[width=\linewidth]{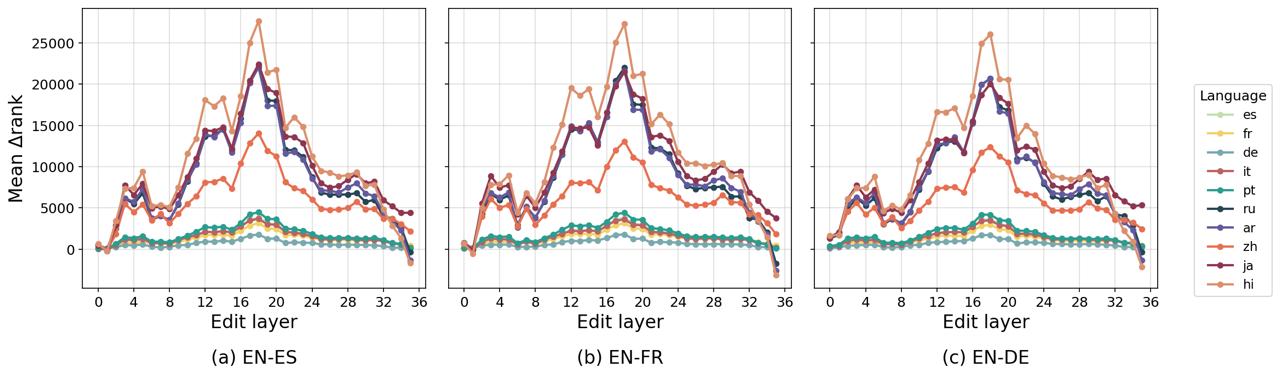}
    \caption{\textbf{Mean $\Delta$rank across edit layers for different target languages for Tiny-Aya.} Curves correspond to target languages, and higher values mean that the FV
moves the correct translation token closer to the top of the next-token distribution. 
}
    \label{fig:mean-rank-aya}
\end{figure*}

\begin{figure*}[t]
    \centering
    \includegraphics[width=\linewidth]{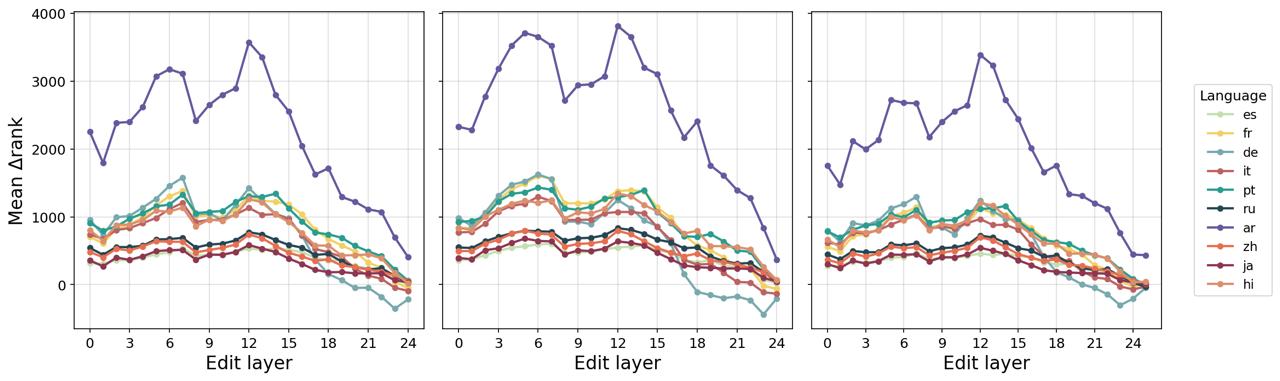}
    \caption{\textbf{Mean $\Delta$rank across edit layers for different target languages for Gemma-2-2B.} Curves correspond to target languages, and higher values mean that the FV
moves the correct translation token closer to the top of the next-token distribution. 
}
    \label{fig:mean-rank-gemma}
\end{figure*}

Figures~\ref{fig:mean-rank-aya},~\ref{fig:mean-rank-gemma}, and~\ref{fig:mean-rank-llama} report the mean $\Delta$rank across edit layers for each target language on Tiny-Aya, Gemma-2-2B, and Llama, respectively, providing a rank-based view of the function vector's effect on the next-token distribution.
\label{sec:llama-three-in-row}

\FloatBarrier
\clearpage

\FloatBarrier

\newpage

\subsection{Ablation and specificity.}
\label{app:ablation}
 The preceding results show that adding the FV promotes correct translations across target languages. We now test the converse: whether removing the FV direction degrades translation, and whether any such degradation is specific to translation.
At a chosen layer  \(\ell\) we project out the component of each residual-stream activation aligned with the FV direction:
\begin{equation}
\tilde{h}_{\ell,i}
= h_{\ell,i} - \alpha \bigl(h_{\ell,i} u^\top\bigr) u,
\qquad
u=\frac{v}{\lVert v\rVert_2}.
\end{equation}, where $v$ is the EN→[X] translation FV and u its unit vector.
Removing the FV direction degrades the rank of the gold translation token consistently across all three FVs (Figure X). Crucially, the degradation extends beyond the extraction direction to other target languages and is concentrated in the same middle-layer band where FV addition produces the largest gains. Addition and ablation thus produce symmetric, cross-lingual effects localized to the same intermediate representations, indicating the FV direction carries translation information the model actively uses rather than an artifact of the addition intervention.
A rank degradation is only informative if it reflects translation-specific computation rather than broad disruption. We therefore evaluate the ablated model on standard NLP benchmarks.

As shown in Table~\ref{tab:ablation-summary}, accuracy changes by at most 1–2 points across all models and tasks, while translation rank degrades substantially. The ablation is therefore largely selective to translation, supporting the interpretation that the FV direction encodes a translation-relevant subspace rather than a general-purpose feature.

\begin{figure*}[t]
    \centering
    \begin{subfigure}[t]{0.32\textwidth}
        \centering
        \includegraphics[width=\linewidth]{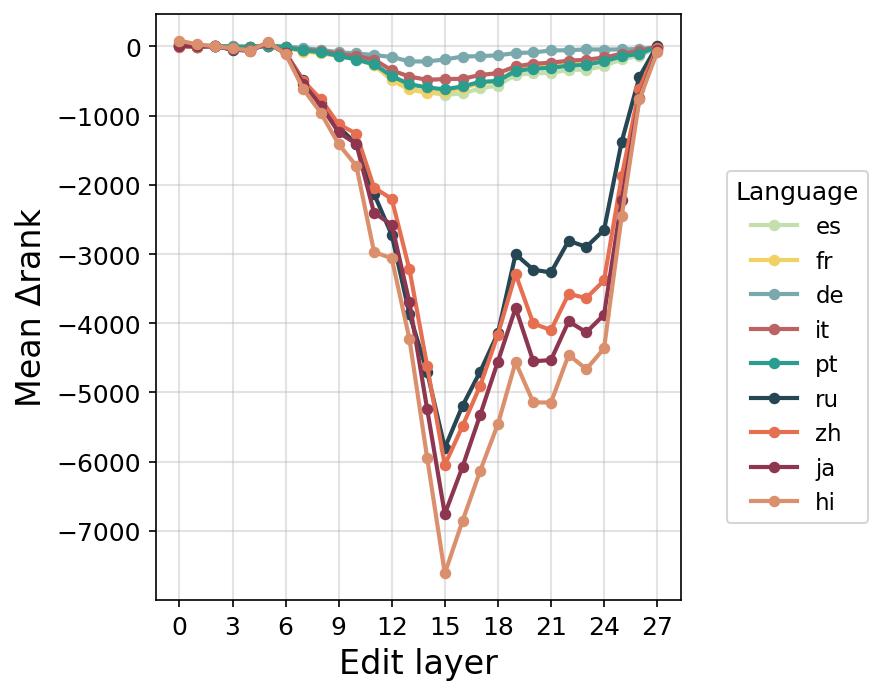}
        \caption{EN-ES}
        \label{fig:img1}
    \end{subfigure}
    \hfill
    \begin{subfigure}[t]{0.32\textwidth}
        \centering
        \includegraphics[width=\linewidth]{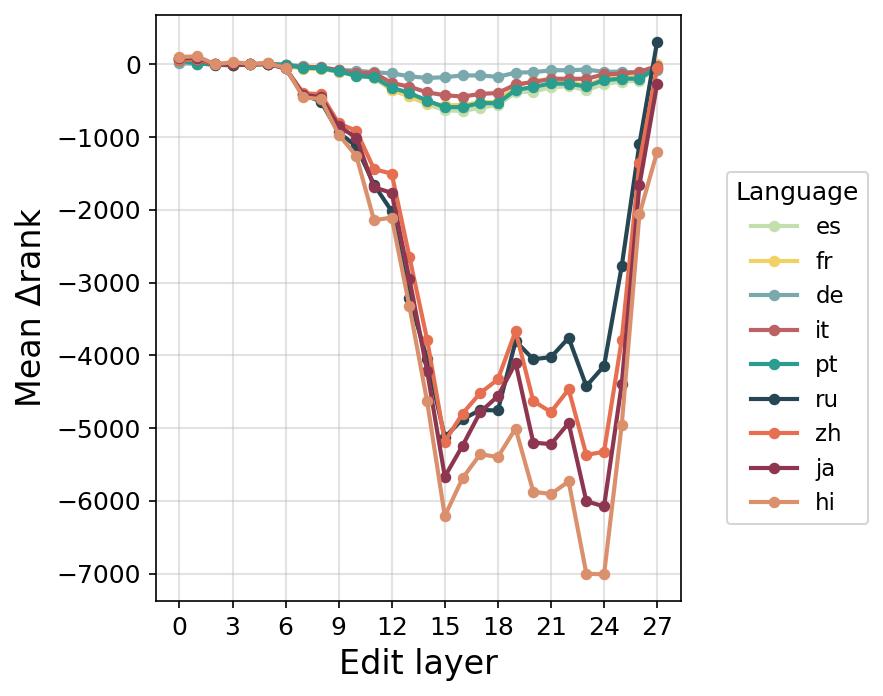}
        \caption{EN-DE}
        \label{fig:img2}
    \end{subfigure}
    \hfill
    \begin{subfigure}[t]{0.32\textwidth}
        \centering
        \includegraphics[width=\linewidth]{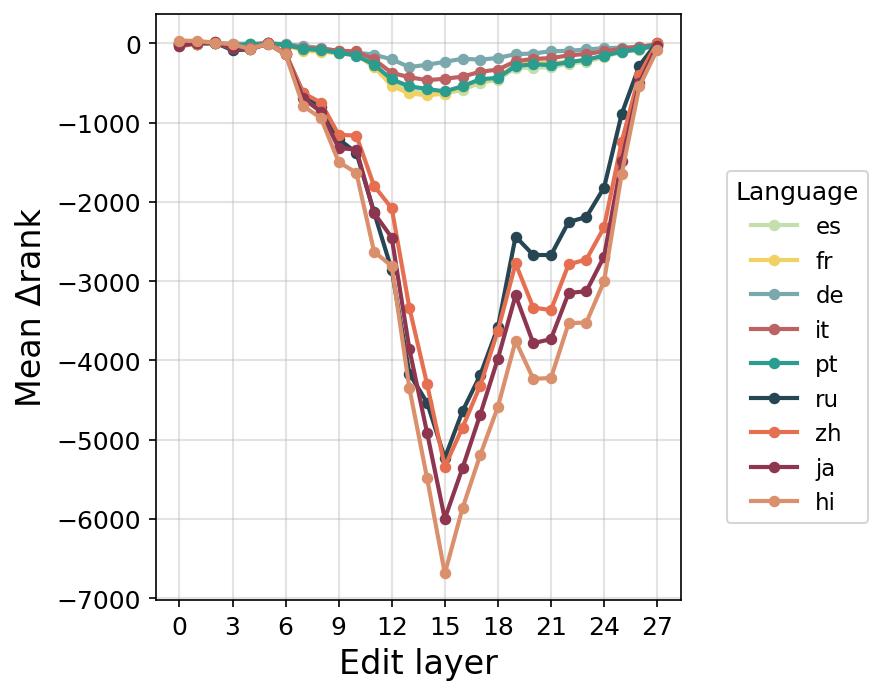}
        \caption{EN-FR}
        \label{fig:img3}
    \end{subfigure}
\caption{\textbf{Mean $\Delta$rank across edit layers for different target languages under FV ablation for Llama-3.2-3B.}
Curves correspond to target languages. More negative values indicate that ablating the FV moves the correct translation token lower in the next-token ranking relative to the clean setting.}
    \label{fig:llama-ablation}
\end{figure*}

\begin{figure*}[t]
    \centering
    \includegraphics[width=\linewidth]{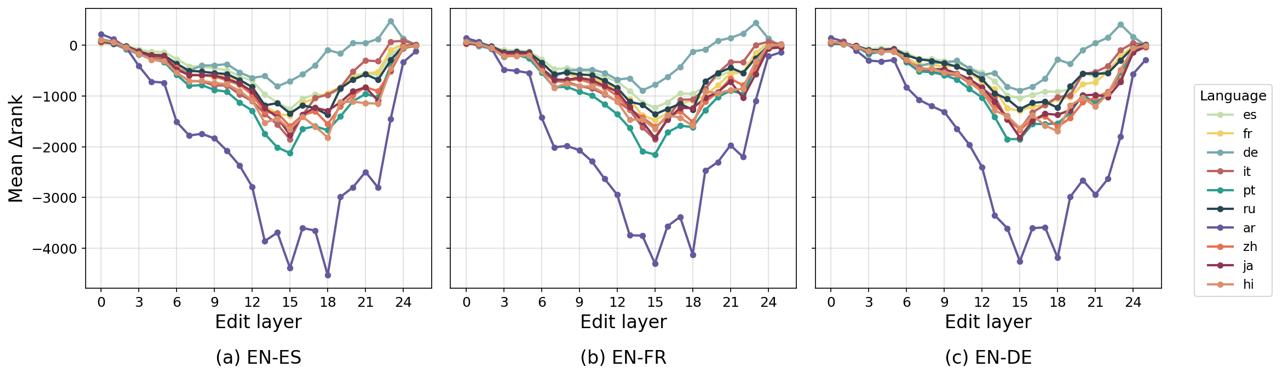}
\caption{\textbf{Mean $\Delta$rank across edit layers for different target languages under FV ablation for Gemma-2-2B.}
Curves correspond to target languages. More negative values indicate that ablating the FV moves the correct translation token lower in the next-token ranking relative to the clean setting.}
    \label{fig:gemma-ablation}
\end{figure*}

\begin{figure*}[t]
    \centering
    \includegraphics[width=\linewidth]{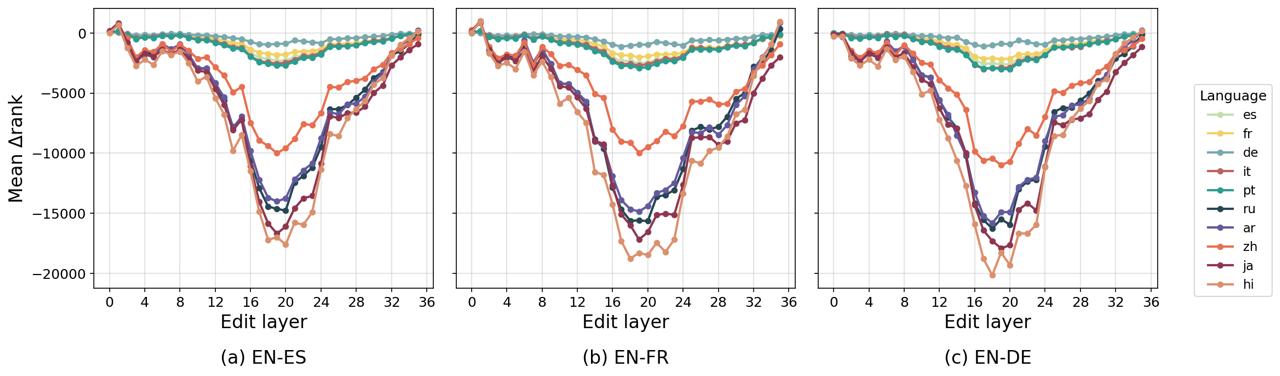}
\caption{\textbf{Mean $\Delta$rank across edit layers for different target languages under FV ablation for Tiny-Aya.}
Curves correspond to target languages. More negative values indicate that ablating the FV moves the correct translation token lower in the next-token ranking relative to the clean setting.}
    \label{fig:aya-ablation}
\end{figure*}

\input{tables/ablation}

%% file: tables/ablation.tex
\definecolor{pagegray}{RGB}{241,241,241}
\definecolor{headergray}{RGB}{235,235,235}
\definecolor{bandpink}{RGB}{217,234,211}
\definecolor{bandblue}{RGB}{211,229,239}
\definecolor{rulegray}{RGB}{60,60,60}

\renewcommand{\arraystretch}{1.18}

\begin{table*}[t]
\centering
\small
\setlength{\tabcolsep}{6pt}
\resizebox{\textwidth}{!}{%
\begin{tabular}{p{1cm}*{11}{>{\centering\arraybackslash}p{1.6cm}}}
\toprule
\rowcolor{headergray}
\textbf{Data} & \textbf{ARC\_EASY} & \textbf{BOOLQ} & \textbf{COPA} &
\textbf{HELLA-}\newline\textbf{SWAG} &
\textbf{OPEN-}\newline\textbf{BOOKQA} &
\textbf{PIQA} & \textbf{SCIQ} & \textbf{WINO-}\newline\textbf{GRANDE} &
\textbf{MMLU\_DE} & \textbf{MMLU\_ES} & \textbf{MMLU\_FR} \\
\midrule

\multicolumn{12}{c}{\cellcolor{bandpink}\textbf{\texttt{Llama-3.2-3B}}} \\
\midrule
Clean   & 0.74 & 0.74 & 0.82 & 0.56 & 0.31 & 0.76 & 0.95 & 0.70 & 0.52 & 0.55 & 0.55 \\
Ablated & 0.74 (+0.00) & 0.74 (+0.00) & 0.83 (+0.01) & 0.56 (+0.00) & 0.31 (+0.00) & 0.76 (+0.00) & 0.95 (+0.00) & 0.69 (-0.01) & 0.53 (+0.01) & 0.55 (+0.00) & 0.54 (-0.01) \\
\midrule

\multicolumn{12}{c}{\cellcolor{bandpink}\textbf{\texttt{Gemma-2-2B}}} \\
\midrule
Clean   & 0.80 & 0.73 & 0.88 & 0.55 & 0.31 & 0.79 & 0.96 & 0.68 & 0.40 & 0.43 & 0.41 \\
Ablated & 0.79 (-0.01) & 0.70 (-0.03) & 0.88 (+0.00) & 0.55 (+0.00) & 0.31 (+0.00) & 0.79 (+0.00) & 0.96 (+0.00) & 0.69 (+0.01) & 0.40 (+0.00) & 0.41 (-0.02) & 0.41 (+0.00) \\
\midrule

\multicolumn{12}{c}{\cellcolor{bandpink}\textbf{\texttt{Tiny Aya}}} \\
\midrule
Clean   & 0.76 & 0.81 & 0.78 & 0.50 & 0.28 & 0.76 & 0.94 & 0.64 & 0.57 & 0.52 & 0.54 \\
Ablated & 0.76 (+0.00) & 0.81 (+0.00) & 0.80 (+0.02) & 0.50 (+0.00) & 0.28 (+0.00) & 0.76 (+0.00) & 0.94 (+0.00) & 0.65 (+0.01) & 0.58 (+0.01) & 0.53 (+0.01) & 0.54 (+0.00) \\
\bottomrule
\end{tabular}}
\caption{Clean vs.\ FV-ablation accuracy across evaluation tasks.}
\label{tab:ablation-summary}
\end{table*}